\documentclass[sponsored]{acmsiggraph}

\usepackage{siunitx}
\usepackage{booktabs}
\usepackage{multirow}
\usepackage{tikz}
\usepackage{subcaption}
\usepackage{url}

\definecolor{mred}{HTML}{A00000}
\definecolor{mgreen}{HTML}{00A000}

\newcommand{\figref}[1]{Figure~\ref{#1}}
\newcommand{\tblref}[1]{Table~\ref{#1}}
\renewcommand{\eqref}[1]{Equation~(\ref{#1})}

\TOGonlineid{138}

\title{Bayesian Identification of Fixations, Saccades, and Smooth Pursuits}

\def\institute{ Perception Engineering Department \\ University of T\"{u}bingen \\ Germany}
\author{
Thiago Santini
\thanks{e-mail: thiago.santini@uni-tuebingen.de}
\\
\institute
\and
Wolfgang Fuhl
\thanks{e-mail: wolfgang.fuhl@uni-tuebingen.de}
\\
\institute
\and
Thomas K\"{u}bler
\thanks{e-mail: thomas.kuebler@uni-tuebingen.de}
\\
\institute
\and
Enkelejda Kasneci
\thanks{e-mail: enkelejda.kasneci@uni-tuebingen.de}
\\
\institute
}

\pdfauthor{Thiago Santini, Wolfgang Fuhl, and Enkelejda Kasneci}

\keywords{smooth pursuit, eye-tracking, probabilistic, model, online,
classification, dynamic stimuli, open-source}

\def\noisePerc{$\approx1.76\%$}
\def\fixCount{18,682}
\def\sacCount{1,296}
\def\purCount{4,143}
\def\datasetCount{24}
\def\ages{$\mu=31.50$, $\sigma=2.59$}
\def\ibdtRec{$\mu=91.42\%$, $\sigma=9.52\%$}
\def\ibdtPrec{$\mu=95.60\%$, $\sigma=5.29\%$}
\def\ibdtSpec{$\mu=95.41\%$, $\sigma=7.02\%$}
\def\ibdtAcc{$\mu=96.95\%$, $\sigma=2.54\%$}
\def\ibdtFpur{$2.18\%$}

\def\ibdtMinPurAcc{$90.57\%$}
\def\ibdtAvgPurAcc{$94.98\%$}
\def\ibdtMaxPurAcc{$98.19\%$}
\def\ibdtPurDRec{0.822}
\def\ibdtPurDSpec{0.984}
\def\ibdtFixDRec{0.986}
\def\ibdtFixDSpec{0.859}
\def\ibdtPurSRec{N/A}
\def\ibdtPurSSpec{N/A}
\def\ibdtFixSRec{0.985}
\def\ibdtFixSSpec{0.977}
\def\ivdtRec{$\mu=87.67\%$, $\sigma=14.73\%$}
\def\ivdtPrec{$\mu=89.57\%$, $\sigma=\phantom{0}8.05\%$}
\def\ivdtSpec{$\mu=92.10\%$, $\sigma=11.21\%$}
\def\ivdtAcc{$\mu=94.65\%$, $\sigma=\phantom{0}4.50\%$}

\begin{document}




\maketitle


\begin{abstract}

Smooth pursuit eye movements provide meaningful insights and information on
subject's behavior and health and may, in particular situations, disturb the
performance of typical fixation/saccade classification algorithms.
Thus, an automatic and efficient algorithm to identify these eye movements is
paramount for eye-tracking research involving dynamic stimuli.
In this paper, we propose the Bayesian Decision Theory Identification (I-BDT)
algorithm, a novel algorithm for ternary classification of eye movements that is
able to reliably separate fixations, saccades, and smooth pursuits in an online
fashion, even for low-resolution eye trackers.
The proposed algorithm is evaluated on four datasets with distinct mixtures of
eye movements, including fixations, saccades, as well as straight and circular
smooth pursuits; data was collected with a sample rate of \SI{30}{\hertz} from
six subjects, totaling \datasetCount{} evaluation datasets.
The algorithm exhibits high and consistent performance across all datasets and
movements relative to a manual annotation by a domain expert
(recall: \ibdtRec{}; precision: \ibdtPrec{}; specificity \ibdtSpec{})
and displays a significant improvement when compared to I-VDT, an
state-of-the-art algorithm
(recall: \ivdtRec{}; precision: \ivdtPrec{}; specificity \ivdtSpec{}).
For the algorithm implementation and annotated datasets, please contact the first
author.

\end{abstract}


\begin{CRcatlist}
  \CRcat{I.5.1}{Computing Methodologies}{Pattern Recognition -- Models}{};
  \CRcat{I.6.4}{Computing Methodologies}{Simulation and Modeling -- Model
  Validation and Analysis}{};
  \CRcat{J.7}{Computer Applications}{Computers in Other Systems -- Real Time}{};
\end{CRcatlist}


\keywordlist




\copyrightspace

\section{Introduction}

The human visual perception involves mainly six types of eye movements:
fixations, saccades, smooth pursuits, optokinetic reflex, vestibulo-ocular
reflex, and vergence~\cite{leigh2015neurology}.
The automatic and correct identification of these eye movements based on the raw
eye-position signal is critical for research and applications involving eye
trackers -- such as cognitive science and medical research, task assistance
(e.g., driving) and marketing applications, and Human Computer Interfaces (HCI).

Initially, eye-tracking research restrained head movements and employed
\emph{static stimuli}, such as images and text.
In this scenario, the only relevant movements considered were \emph{fixations}
(in which the eyes are relatively still) and \emph{saccades} (rapid transitions
from one fixation point to another); thus, early algorithms for the automatic
classification of eye movements focused on segregating only between these two
movements.
Nowadays, there is an increasing interest in using \emph{dynamic stimuli} (e.g.,
video clips)~\cite{larsson2015detection}.
With dynamic stimuli, it is often the case that an object of interest moves
through the subject's field of view.
As a result, the subject tracks this object to keep it within the fovea,
producing a fluent eye motion -- which we denominate a \emph{smooth pursuit}.

The presence of smooth pursuits disturbs the performance of established
fixation/saccade classification algorithms since these pursuits end up spread
over the two classification classes.
Moreover, they also provide valuable information on subject's health and
behavior; for instance, smooth pursuit impairment and dysfunction have been
linked to mental illnesses, such as schizophrenia~\cite{o2008smooth} and
Alzheimer's disease~\cite{fletcher1988smooth}.
Thus, an automatic and efficient algorithm to distinguish between fixations,
saccades, and smooth pursuits is paramount for eye-tracking research involving
dynamic stimuli.
Furthermore, some of the possible applications must be on the form of embedded
systems (e.g., driving assistance) and impose real-time, processing, and energy
consumption constrains on the eye-tracking system.
To meet these constraints, typically eye trackers with a lower sample rate are
used.  Consequently, such an algorithm must not only work in real-time, but also
be able to deal with the low resolution arising from such eye trackers.

In this paper, we propose a novel algorithm for ternary classification of
oculomotor events. Our main contributions are:
\begin{itemize}
	\item We propose the Bayesian Decision Theory Identification (I-BDT)
		algorithm to identify fixations, saccades, and smooth pursuits in
		real-time for low-resolution eye trackers. Additionally, the algorithm
		operates directly on the eye-position signal and, thus, requires no
		calibration.
	\item The proposed algorithm is evaluated relative to manual annotation by a
		domain expert, and performance is measured in terms of recall, precision,
		specificity, and accuracy; on average, the proposed algorithm scores above
		$90\%$ on all metrics.
	\item I-BDT's performance is compared to that of an state-of-the-art
		algorithm (Velocity and Dispersion Threshold Identification), showing a
		significant improvement in terms of average score and variability.
	\item Additionally, we provide a \emph{MATLAB} implementation for the I-BDT
		algorithm as well as the annotated datasets used for evaluation. Please
		contact the first author for these.
\end{itemize}

\section{Related Work}

In 1991, \cite{sauter1991analysis} proposed using a Kalman filter coupled with a
$\chi^2$-test to separate saccades from other eye movements.  This approach was
later extended as the Attention Focus Kalman Filter (AFKF)
by~\cite{komogortsev2007kalman}, using velocity and temporal thresholds to
separate fixations from smooth pursuits.
Similarly, several methods use a simple \emph{velocity threshold} to isolate
saccades, followed by a second step to distinguish between fixations and smooth
pursuits.
These are typically identified by a name following the pattern \emph{I-V*}.
\cite{komogortsev2013automated} proposed to distinguish between the remaining
movements through a \emph{second velocity} threshold (Velocity and Velocity
Threshold Identification (I-VVT)) and through a \emph{dispersion} threshold
combined with a temporal window (Velocity and Dispersion Threshold
Identification (I-VDT)).
\cite{berg2009free} proposed analyzing the ratio between first and second
principal components to identify smooth pursuits (Principal
Component Analysis Identification (I-PCA)) on the intuition that fixations
would have a ratio close to one.
\cite{lopez2009off} started a subgroup that uses the \emph{movement pattern} to
identify smooth pursuits, hence the common prefix \emph{I-VMP};
\cite{lopez2009off} used the standard deviation of the movement directions in a
time window to isolate fixations (Velocity Movement Pattern Standard
Deviation Identification (I-VMPStd)).
\cite{larsson2010} used a Rayleigh test to identify smooth pursuits by rejecting
the hypothesis of uniformity of inter-sample vectors around the unit circle
(Velocity Movement Pattern Rayleigh Identification (I-VMPRay)); more
recently, this algorithm was extended with four different spatial features
(dispersion, consistent direction, positional displacement, and spatial range)
in~\cite{larsson2015detection}.

\cite{tafaj2012bayesian} used a Bayesian Mixture Model based on the Euclidean
distance between sequential points to discern fixations from saccades, which was
later extended in~\cite{kasneci2013towards} with a principal component analysis
similar to I-PCA to identify smooth pursuits. This method is called the Bayesian
Mixture Model Identification (I-BMM).
\cite{vidal2012detection} defined a set of shape features, whose expected range
is derived from training data.  A k-nearest neighbors classifier ($k=3$) is then
used to isolate smooth pursuits from other movements.

\begin{figure}[ht]
	\centering
	\begin{tikzpicture}
		\draw [ultra thick,opacity=0.6] (.34\columnwidth,.5\columnwidth)
		circle [radius=.33\columnwidth];
		\draw [ultra thick,opacity=0.6,dashed] (.65\columnwidth,.5\columnwidth)
		circle [radius=.33\columnwidth];
		\node [text width=.33\columnwidth,align=center]
		at (.165\columnwidth,.5\columnwidth){
			I-VDT\\
			I-VVT\\
			I-PCA\\
			I-VMPStd\\
		};
		\node [text width=.33\columnwidth,align=center]
		at (.495\columnwidth,.5\columnwidth){
			\cite{larsson2015detection}\\
			I-BMM\\
			I-VMPRay\\
			AFKF\\
		};
		\node [text width=.33\columnwidth,align=center]
		at (.825\columnwidth,.5\columnwidth){
			\color{mgreen}
			\emph{\textbf{I-BDT}}\\
			\color{black}
			\cite{vidal2012detection}
		};
		\node [text width=.33\columnwidth,align=center]
		at (.33\columnwidth,.85\columnwidth){
			Threshold-based
		};
		\node [text width=.33\columnwidth,align=center]
		at (.66\columnwidth,.85\columnwidth){
			Probabilistic
		};
	\end{tikzpicture}
	\caption{
		Algorithms for the automatic identification of smooth pursuits according
		to a broad classification based on their underlying mechanisms. The
		algorithm proposed on this work (I-BDT) falls within the probabilistic
		group.
	}
	\label{fig:groups}
\end{figure}
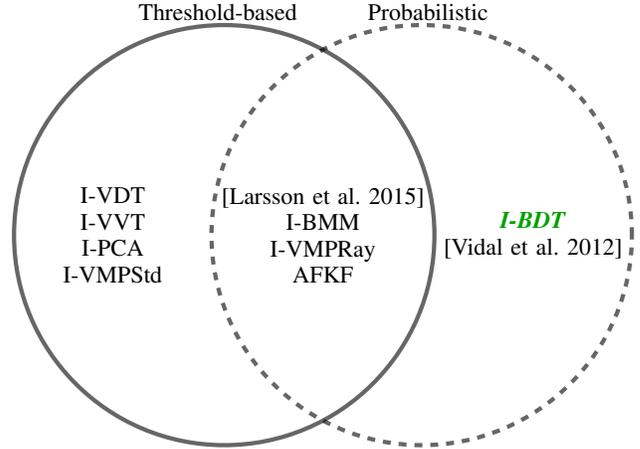

As can be seem in \figref{fig:groups}, these methods for automatic
classification of eye movements fall mainly into two classes:
\emph{threshold-based} and \emph{probabilistic} methods.
While threshold-based algorithms tend to be simpler to implement, their major
drawback is that they usually depend on the eye movements being clearly
discernible from each other.
On the other hand, probabilistic methods work based on softer decision rules in
the form of probabilities, making them more flexible.
Hybrid methods combine insights from physiological limits to define clear
thresholds (e.g., only during saccades the eyes reach velocities above
\SI{100}{\degree\second}~\cite{meyer1985upper}) with a probabilistic approach
in other cases.
I-BDT, the method proposed in this paper, falls into the probabilistic
group.

Furthermore, most previous work has focused on eye trackers with high sampling
rates (i.e., above \SI{250}{\hertz}).
However, in dynamic scenarios where a non-intrusive head-mounted eye tracker is
required (e.g., driving assistance), such high sampling rates are not available.
Currently, mostly head-mounted eye trackers present an upper limit of
\SI{60}{\hertz} for binocular tracking (e.g., Dikablis Pro, SMI Glasses 2, ASL
H7 Optics, Tobii Pro Glasses 2).
The exception is SR Research's EyeLink II, which has a binocular sampling rate
of \SI{500}{\hertz}.
Despite its clear advantage in temporal resolution, this eye tracker is
rather intrusive, occupying a large part of the subject's field of view;
for comparison, EyeLink II's eye cameras measure each approximately
\SI{5x5x1}{\cm} while Dikablis Pro's eye cameras measure approximately only
\SI{2.5x2x1}{\cm}, resulting in a volume difference of five times.

\section{Bayesian Decision Theory Identification}

\subsection{Problem Statement}

Let $ S = \{ s_i | 1 \le i \le N \} $ be a set of $N$ temporally ordered tuples,
each containing two-dimensional pupil position estimates $(x_i, y_i)$ and a
timestamp $(t_i)$ generated by an eye tracker (i.e., an eye-tracker protocol).
The problem, thus, is to classify all periods between two subsequent tuples
according to the set of possible events $ E = \{ fix, sac, pur \} $, where
$fix$, $sac$, and $pur$ stand respectively for fixation, saccade, and smooth
pursuit.

\subsection{Model}

In this paper, we propose a Bayesian decision theory approach to solve the
stated problem based on a pair of features derived from $S$.
In other words, given some data $D$, we are interested in defining the
\emph{likelihoods} $p(D|e)$ and \emph{priors} $p(e)$ for each event $e \in E$ in
order to calculate the \emph{posteriors} $p(e|D)$ of these events.
Following the naming convention from \cite{komogortsev2013automated} and
\cite{salvucci2000identifying}, we will hereby refer to this method as the
Bayesian Decision Theory Identification (I-BDT) algorithm.

The first feature derived from $S$ is the estimated eye speed ($v_i$) between
two subsequent tuples, defined as
\begin{equation}
	\label{eq:v}
	v_i =
	{
		\sqrt{
		{\Delta x_i}^2
		+
		{\Delta y_i}^2
	}
	\over
		\Delta t_i
	}
\end{equation}
where
$\Delta x_i = x_i - x_{i-1}$,
$\Delta y_i = y_i - y_{i-1}$, and
$\Delta t_i = t_i - t_{i-1}$.

\begin{figure}[ht]
  \centering
  \includegraphics[width=\columnwidth]{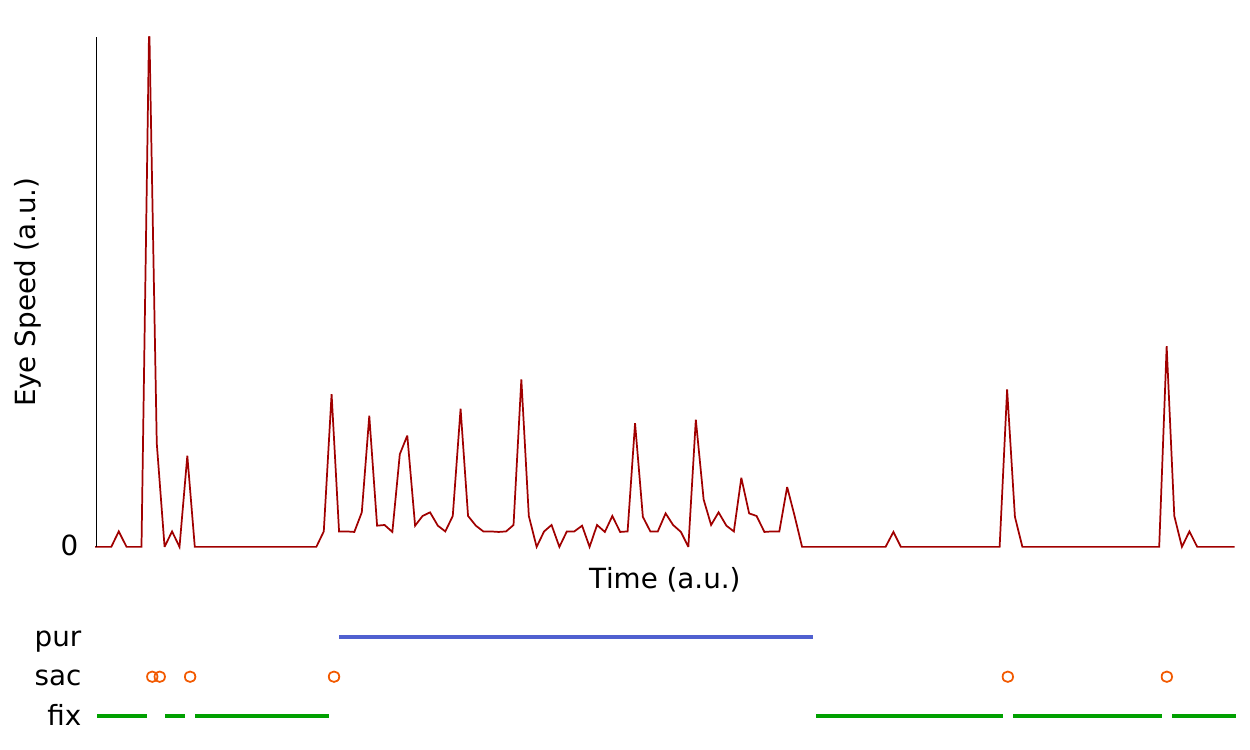}
  \caption{
	  Eye speed compared to manual classification by a domain expert. Fixations
	  (fix) tend to be mostly still, with only few deviations due to
		micro eye movements and measurement noise, while saccades (sac) result in brief
	  spikes in the eye speed signal. On the contrary, smooth pursuits (pur)
	  show a distinct speed pattern during a longer period of time.
  }
  \label{fig:feature2}
\end{figure}

The second derived feature is the movement ratio $r_i$ over the window $ W_i =
\{ v_j | i-N_w < j \le i \} $ of the latest $N_w$ tuples.
For simplicity, we will define it as the amount of non-zero eye speed estimates
relative to the window size, conveying the idea that the more movement in the
window, the more likely a smooth pursuit is; thus,
\begin{equation}
	\label{eq:r}
	r_i =
	{
		{ {1} \over { N_w } }
		\sum\limits_{v_j \in W_i} [ v_j > 0 ]
	} =
	{
		{ {1} \over { N_w } }
		\sum([ W_i > 0 ])
	}
\end{equation}
where $[X]$ is the Iverson bracket notation~\cite{knuth1992two} given by
\begin{equation*}
	[X] =
	\begin{cases}
		1 & \mbox{if $X$ is true;}\\
		0 & \mbox{otherwise.}\\
	\end{cases}
\end{equation*}
It is paramount to note that this feature's definition is heavily dependent on
the eye tracker used to record the data and its temporal and spatial resolution; zero speed may not
be an appropriate representation for fixations.
Nevertheless, the intuition behind this feature is that fixations exhibit little
continuous movement, saccades are brief and usually separated by fixations, and
smooth pursuits tend to exhibit continuous movement during larger periods of
time (see~\figref{fig:feature2}).
Therefore, $r_i$ should be a good smooth pursuit indicator if an adequate window
size is chosen; this time window should be large enough to encompass the maximum
saccade duration, otherwise misclassification of saccades as pursuits may be
exacerbated.
In our model, we use this feature directly as the smooth pursuit likelihood,
i.e.,
\begin{equation}
	p(r_i | pur) = r_i.
\end{equation}

Once a smooth pursuit has started, it tends to continue for
an arbitrary period; thus, this should be reflected on one's belief before any
evidence is taken into account. For this reason, we model the smooth pursuit
prior as the mean of previous smooth pursuit likelihoods (i.e., the set
$ L_i = \{ p(r_j | pur) | i-N_w < j < i \}$) such that
\begin{equation}
	p(pur) =
	{
		{ {1} \over { N_w - 1 } }
		\sum\limits_{ p(r_j | pur) \in L_i} p(r_j | pur)
	}.
\end{equation}

Naturally, the joint probability of priors must sum to one. With no further
evidence, we do not have reason to believe either fixations or saccades
are more probable, and, thus, we divide the remaining joint prior probability
equally between these movements such that
\begin{equation}
	p(fix) =
	p(sac) =
	{
		{1 - p(pur)} \over {2}
	}.
\end{equation}
It is worth noticing that if information on the task being performed by the
subject is available, one could improve these priors based on the duration of
the current event. For instance, imagine a task characterized by fixations with
a relatively constant duration: after a first fixation is found, the
following events are likely to be fixations until the average fixation duration
is reached. At this point the next event becomes less and less probable to
be a fixation. Such behaviour could be taken into account by adjusting the priors.

\begin{figure}[ht]
  \centering
  \includegraphics[width=\columnwidth]{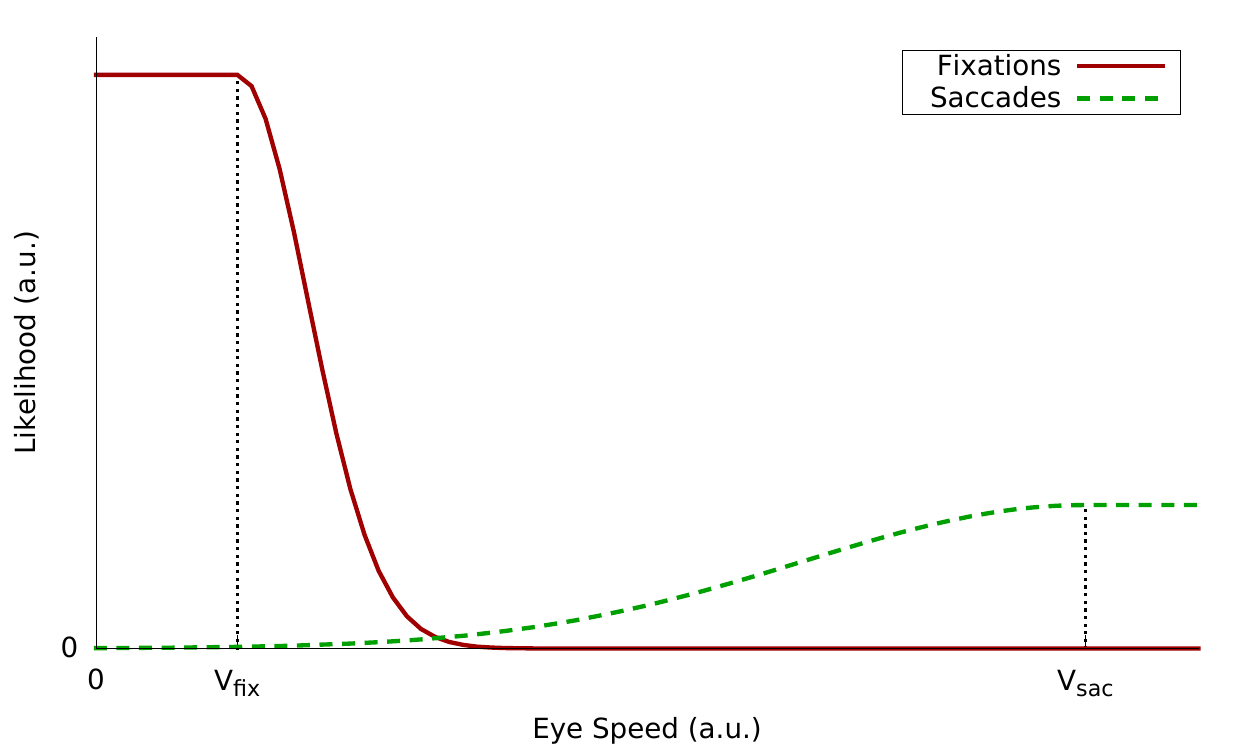}
  \caption{Resulting fixational and saccadic likelihoods based on the eye speed
  feature ($v_i$).}
  \label{fig:likelihood}
\end{figure}

The fixational and saccadic likelihoods are deemed to be dependent only on
the current eye speed ($v_i$) feature.
This feature can be used to reliably separate high-speed saccades from other
events as it has been shown that no other event can reach a velocity higher than
$V_{sac}$, estimated to be around \SI{100}{\degree/\second}~\cite{meyer1985upper}.
However, the speed spectra of different eye movements overlap for lower velocities.
Nonetheless, it is intuitive that velocities closer to zero are more likely to stem
from fixations while velocities closer to $V_{sac}$ are more likely to stem from
saccades.
In fact, \cite{tafaj2012bayesian} have shown that saccades and fixations can be
represented by a mixture model of two Gaussian distributions based on the
distance between sequential points -- one Gaussian generating fixations, and
another one generating saccades.
Therefore, we assume the eye speed feature to also be generated by two such
Gaussian distributions.
Intuitively, saccade likelihood should be at its maximum for speeds above
$V_{sac}$.
Ideally, fixations would exhibit zero speed; however, as they typically include
small movements, such as microsaccades and tremors, there is a threshold speed
$V_{fix}$ that encompasses these combination of movements.
Thus, fixation likelihood should be at its maximum for speeds below $V_{fix}$.
In the interval between these thresholds, we assume the likelihood to be
generated by two Gaussian\footnote{
	Denoted as
	\begin{equation*}
	N(x | \mu, \sigma) =
	\frac{1}{{\sigma \sqrt {2\pi } }}e^{{{ - \left( {x - \mu } \right)^2 }
	\mathord{\left/ {\vphantom {{ - \left( {x - \mu } \right)^2 } {2\sigma ^2
	}}} \right. \kern-\nulldelimiterspace} {2\sigma ^2 }}}
	\end{equation*}
} distributions, one centered around $V_{fix}$ and the other around
$V_{sac}$ (see~\figref{fig:likelihood}).
Thus,
\begin{equation}
	\label{eq:nfix}
	p(v_i | fix) =
	\begin{cases}
		N(V_{fix}|V_{fix},\sigma_{fix}) & \mbox{if }  v_i < V_{fix} \\
		N(v_{i}\quad|V_{fix},\sigma_{fix}) & \mbox{if }  v_i \ge V_{fix} \\
	\end{cases},
\end{equation}
and
\begin{equation}
	\label{eq:nsac}
	p(v_i | sac) =
	\begin{cases}
		N(v_i\quad|V_{sac},\sigma_{sac}) & \mbox{if }  v_i < V_{sac} \\
		N(V_{sac}|V_{sac},\sigma_{sac}) & \mbox{if }  v_i \ge V_{sac} \\
	\end{cases}.
\end{equation}

Having defined the priors and likelihoods for all events, we can calculate the
posterior for each event $e \in E$ given the data $D = \{ v_i, r_i \}$ using
Bayes' Theorem; thus,
\begin{equation}
	p(e|D) =
	{
		{p(e)p(D|e)}
		\over
		{p(D)}
	},
\end{equation}
and the period is classified as the event with highest posterior probability.
Here, $p(D)$ is merely a scaling factor that guarantees that the sum of the
posterior probabilities sum to one.

\section{Experimental Setup}

\subsection{Dataset}

To evaluate the proposed algorithm, we designed an experiment to cover a wide
range of induced as well as natural eye movements.
The induced movements are characterized in \tblref{tbl:mov}.
\begin{table}[h]
	\centering
	\begin{tabular}{ccc}
	\toprule
	\textbf{Movement} & \textbf{Amplitude$^a$/Radius$^b$ (\si{\degree})} &
	\textbf{Velocity (\si{\degree/\second})} \\
	\midrule
	Saccade$^a$ & 6, 11, 14 & ---\\
	Straight Pursuit$^a$ & 6, 12, 22, 28 & 10, 20, 30\\
	Circular Pursuit$^b$ & 6, 8, 14 & 18, 25, 44 \\
	\bottomrule
	\end{tabular}
	\caption{
		Induced movements used within the experiment. Degrees are expressed in
		terms of visual angle. Straight pursuit amplitudes and velocities were
		combined such that their durations were within 0.4 and 2 seconds to
		account for subject latency while keeping pursuit duration realistic.
		Circular pursuits were conducted at a constant angular velocity of 180\si{\degree}/s.
		Pursuits were separated from other movements
		by one second fixations.  Saccades were separated from each other by
		fixations of 0.75 seconds.
		The directions of the movements were chosen randomly and differ per
		subject.
	}
	\label{tbl:mov}
\end{table}

\begin{figure}[ht]
	\centering
	\begin{subfigure}[b]{0.4\columnwidth}
		\includegraphics[width=\columnwidth]{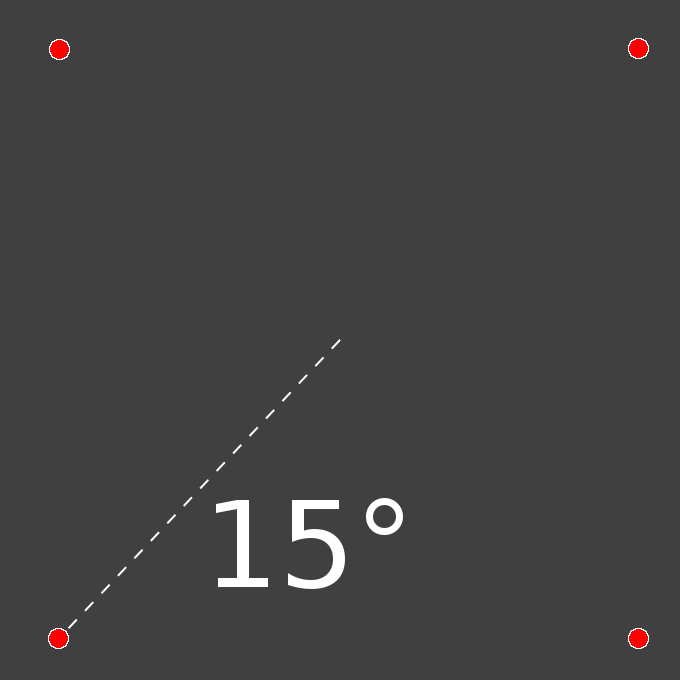}
		\caption{}
		\label{fig:free}
	\end{subfigure}
	\begin{subfigure}[b]{0.4\columnwidth}
		\includegraphics[width=\columnwidth]{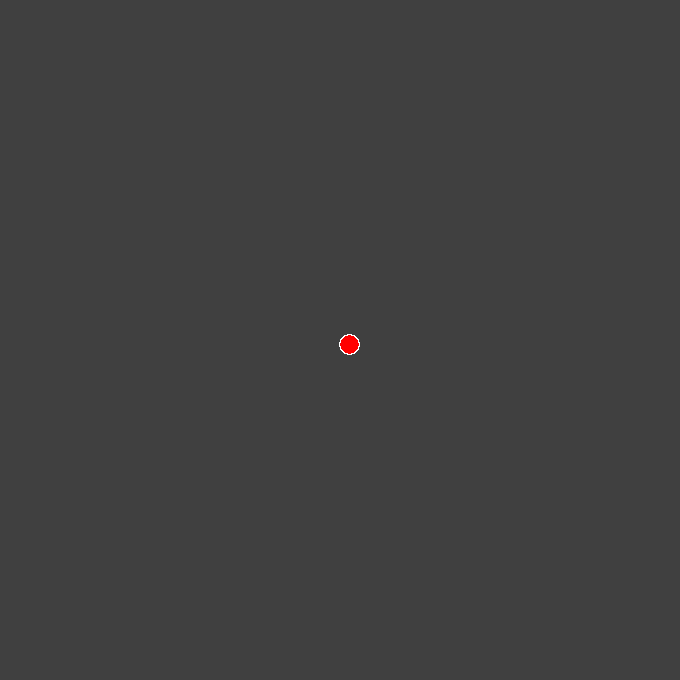}
		\caption{}
		\label{fig:center}
	\end{subfigure}
	\caption{
		Common stimuli at the beginning of each dataset.
	}
\end{figure}

Prior to the recording, each user was shown a tutorial with detailed on-screen
instructions and examples of movements for each class in \tblref{tbl:mov}.
Four datasets were recorded per subject, and all datasets had a common
beginning: first, four dots were shown at \SI{15}{\degree} of visual angle
diagonally from the screen center for five seconds (\figref{fig:free}); subjects
were instructed to look at these stimuli at will.
During this period natural saccades and fixations are collected; saccades of
$\approx$ 20 and 30\si{\degree} of visual angle were expected, separated by
fixations of arbitrary duration.
Afterwards, a single dot appeared at the screen center for two seconds
(\figref{fig:center}); subjects were instructed to focus on and follow this
target.
The subsequent movements differ per dataset and are listed in
\tblref{tbl:datasets}.
\begin{table}[h]
	\small
	\centering
	\begin{tabular}{ll}
	\toprule
	\textbf{Dataset} & \textbf{Movements} \\
	\midrule
	I & Fixations, saccades, and all possible straight pursuits.\\
	II & Fixations and saccades. No pursuits.\\
	III & Fixations, saccades, and all circular pursuits.\\
	IV & Fixations, saccades, straight and circular
	pursuits.\\
	\bottomrule
	\end{tabular}
	\caption{
		Movements distribution per dataset.
	}
	\label{tbl:datasets}
\end{table}

\begin{figure}[ht]
  \centering
  \includegraphics[width=\columnwidth]{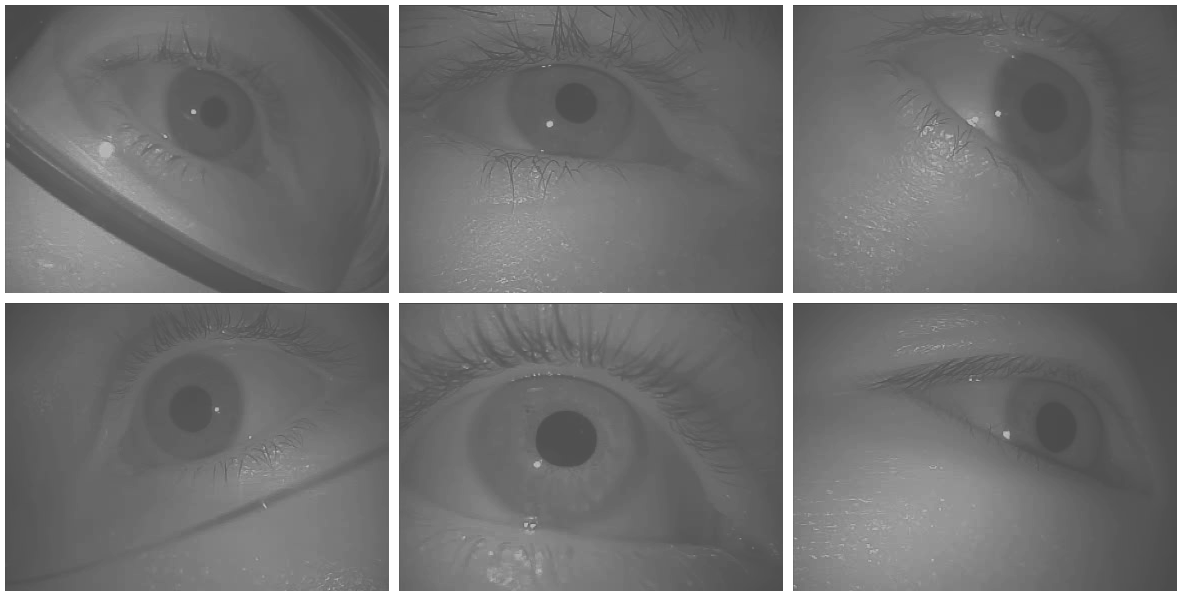}
  \caption{Example of eye location relative to the eye tracker during
  experiments. Note the distinct proximities, positions, and rotations.}
  \label{fig:eyes}
\end{figure}

Targets were red dots (with a width of \SI{1}{\degree} of visual angle) on a
dark gray background displayed using \emph{MATLAB} (r2013a) and the
\emph{Psychtoolbox} (3.0.12)~\cite{kleiner2007s} on a Windows 64-bit machine.
Subjects' heads were supported by a chin rest at a distance of
\SI{300}{\milli\meter} from a \emph{Samsung SyncMaster 2443BW}\footnote{
	Width: \SI{520}{\milli\meter}. Height: \SI{320}{\milli\meter}. Resolution:
	1920x1200 pixels. Screen refresh rate: \SI{60}{\hertz}. Luminance:
	\SI{0.08}{\candela/\square\meter}.
} color display unit.
Ocular dominance was determined using the Miles test, and data was collected
only from the dominant eye using a \emph{Dikablis Pro} eye tracker
(eye images of 384x288 pixels with a 30 Hz sampling rate) and \emph{EyeRec}
(1.2.2) running the \emph{ExCuSe}~\cite{fuhl2015excuse} pupil detection
algorithm on a distinct Windows 64-bit machine.
To avoid gaze estimation noise and calibration requirements, we use the pupil
position signal as input; as such, no calibration step was performed.
An unjittering function was applied to this input prior to processing to remove
obvious jitter artifacts (e.g., one sample spikes~\cite{stampe1993heuristic}).
Six adult subjects (age: \ages{} years; 4 males, 2 females) took part
in the experiment.
Eye location relative to the eye tracker varied greatly between subjects to
exacerbate differences in the input signal and stress the algorithm
(see~\figref{fig:eyes}).
Two of the subjects wore corrective glasses for myopia (-13 dpt and 1.5 dpt).

\subsection{Baseline and Metrics}

The collected data was manually classified by one domain expert in order to
identify data that is not coherent with the stimulus information, e.g., because
the subject did not follow the stimulus as instructed. This manual
classification was used as the ground truth.

Nonetheless, it is worth noticing that the manual classification is a subjective
task, especially for data with a temporal resolution where a measurement period
may contain a mixture of the end of a saccade and the beginning of a fixation.
For this reason, we provide our annotated dataset openly to allow for review and
potential improvements.
Initial corrective saccades during pursuit onset were classified as saccades,
while catch-up saccades during pursuit were classified as smooth pursuits.
Fixation classifications encompass small eye tracker noise, drift and
microsaccades.
Blinks, partial pupil occlusions, and pupil detection failures were marked as
noise and are ignored for performance evaluation; these represent \noisePerc{}
of samples.

Overall, \fixCount{} fixations, \sacCount{} saccades, and \purCount{} smooth
pursuits were classified.
Performance is measured through four metrics per movement class, namely:
recall~$\big({{TP} \over {TP+FN} }\big)$,
precision~$\big({{TP} \over {TP+FP} }\big)$,
specificity~$\big({{TN} \over {TN+FP} }\big)$, and
accuracy~$\big({{TP+TN} \over {TP+FP+TN+FN} }\big)$,
where $TP$, $FP$, $TN$, and $FN$ stand for True Positive, False Positive, True
Negative, and False Negative, respectively.
Moreover, we compare the performance of the proposed
algorithm to that of the I-VDT algorithm as implemented
by~\cite{komogortsev2013automated,komogortsev2010standardization}; I-VDT was
chosen as it can be easily adapted to perform online classification on
low-resolution eye trackers, and because it has been shown to exhibit a
competitive performance with smaller variability relative to other
algorithms~\cite{gyllensten2014,komogortsev2013automated}.
Additionally, we also provide \emph{Cohen's Kappa}~\cite{galar2011overview}
values for the overall classification agreement between the algorithms and the
domain expert to account for agreement merely due to chance.

\subsection{Algorithm's Parameters}

\textbf{I-BDT:}
We have chosen a window size to fit 1.5 times the maximum saccade duration
(\SI{80}{\milli\second}~\cite{holmqvist2011eye}). This value was chosen to fill
the minimum size requirement while keeping the window size to a minimum, thus
minimizing the duration of the pursuit detection onset.
For each subject-dataset pair, the Gaussian distributions parameters are derived from
an approximately \SI{15}{\second} of data to demonstrate an online training
procedure.
Initially, the Expectation-Maximization algorithm was used to derive a mixture
of two Gaussian distributions based on speed samples from this period (with the
smallest positive scalar supported by the platform added to the estimated
covariance matrices to ensure they were positive definite).
The parameters of the Gaussian distribution with the highest mean are used as
parameters for saccades in~\eqref{eq:nsac}.
However, due to the low resolution of the eye tracker, the Gaussian distribution
with the smaller mean is heavily biased towards zero and does not describe
fixations adequately;
we chose instead to derive the parameters for~\eqref{eq:nfix} based on the
inherent eye tracker resolution: the minimum dispersion between two samples
larger than zero divided by the inter-sample period was taken as $V_{fix}$, and
$\sigma_{fix}$ was set to ${{2}\over{3}}V_{fix}$ such that $\approx99.7\%$ of
the distribution values lie within the interval $[0,2 V_{fix}]$.
Furthermore, this low resolution also leads to speed samples with null value
during slow smooth pursuits; thus, we have redefined~\eqref{eq:r} as
\begin{equation}
	r_i =
	{
		{ {1} \over { N_w } }
		\sum(smooth( [\, 0 < W_i < V_{sac} \, ]))
	}
\end{equation}
where the $smooth$ function applies the following logical substitutions over the
entire temporal window
\begin{equation*}
	\begin{cases}
		1x1\phantom{1} \to 111 & \mbox{\small always} \\
		1xx1 \to 1111 & \mbox{\small if sample $i-1$ was classified as a smooth pursuit}\\
	\end{cases},
\end{equation*}
with $x$ representing a \emph{don't care} term.
In other words, $r_i$ tolerates a single isolated null speed sample if not
currently in a smooth pursuit; otherwise, it is more lenient and tolerates up to
two isolated null speed samples.
This redefinition implies the temporal window must include at least four samples.

\textbf{I-VDT:}
In order to get I-VDT's optimal performance, we give it an advantage by
defining pareto-optimal thresholds that maximize Z1 scores based on the
ground truth.
First, the Z1 score for saccade classification is evaluated for all the
inter-sample velocities that can be derived from the eye-tracker protocol; the
velocity that maximizes this score is chosen as the \emph{velocity threshold}.
Second, the minimum fixation duration is derived from the ground truth and is
used as the \emph{temporal window size threshold} (generally around
\SI{100}{\ms}).
Lastly, fixing the previously defined thresholds, the Z1 score for pursuit
classification is evaluated for all the inter-sample dispersions that can be
derived from the eye-tracker protocol; the dispersion that maximizes this score
is chosen as the \emph{dispersion threshold}. If the ground truth contains no
pursuits, the Z1 score for fixation classification is used instead.

\section{Experimental Results}

We start by looking at an overview that encompasses all datasets and movements
to show the overall performance of the proposed algorithm. Afterwards,
we analyze our results for separate movements and datasets to provide a
comprehensive understanding on the behavior of the I-BDT algorithm.
Results are reported using \emph{boxplots}: a box is drawn around the region
between the first and third quartiles, with a horizontal line at the median
value, and the whiskers extend to the minimum and maximum values.
Ideally, the value for all metrics should be as close to one as possible.

\subsection{Overall Results}

\begin{figure}[ht]
	\centering
	\includegraphics[width=\columnwidth]{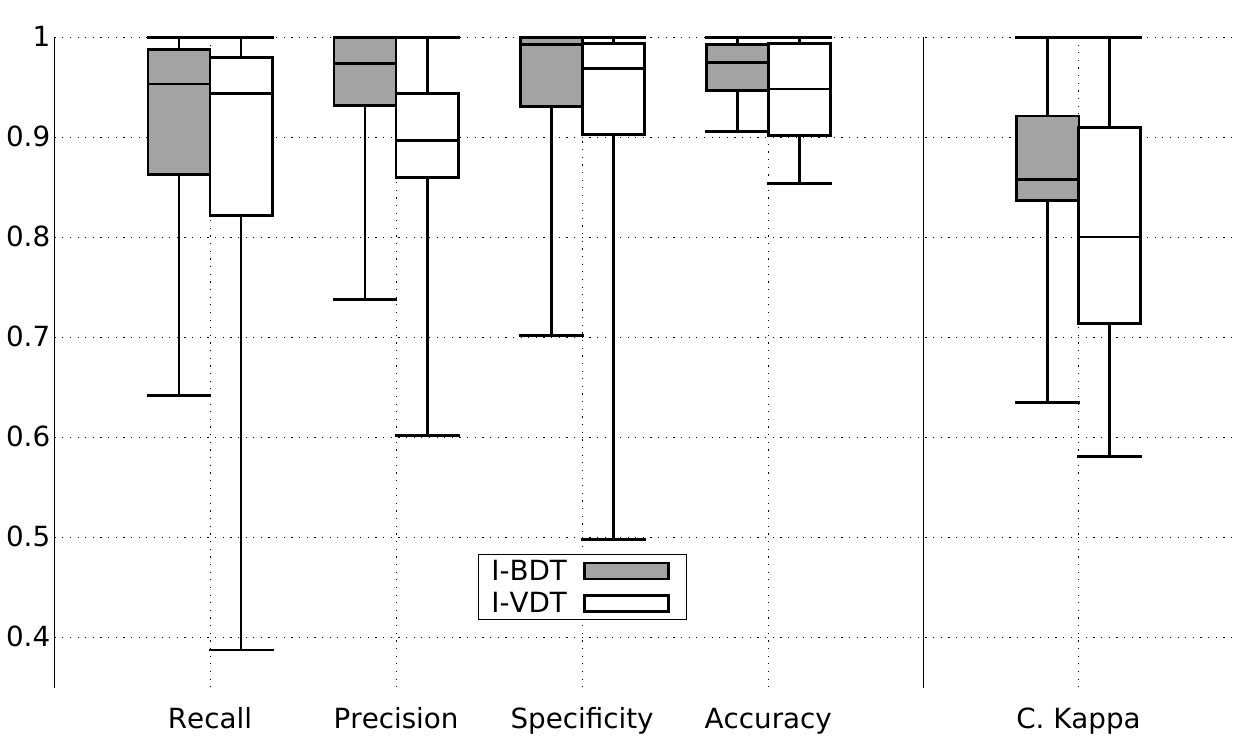}
	\caption{
		Overall algorithm performance. Recall, precision, specificity, and
		accuracy per dataset per subject per movement class
		($n=4\times6\times3=72$).
		Cohen's kappa per dataset per subject ($n=4\times6=24$).
	}
	\label{fig:overview}
\end{figure}

\tblref{tbl:overview} and \figref{fig:overview} indicate the high performance
of the I-BDT algorithm. It is clear that not only I-BDT presents better scores
throughout all metrics relative to I-VDT, it also exhibits less variability.
Moreover, the high Cohen's kappa score indicates that the inter-rater agreement
between expert and algorithm was not due to chance.
Note that, in its current form, the algorithm seems to favor precision instead
of recall;
this is true for smooth pursuits (which can sometimes be misclassified as
fixations, specially during onset) and saccades (which are rarely misclassified
as smooth pursuits);
however, fixations are very seldom misclassified but tend to encompass other
movements in its class more often.
\begin{table}[ht]
	\small
	\centering
	\begin{tabular}{ccc}
	\toprule
	& \textbf{I-BDT} & \textbf{I-VDT} \\
	\midrule
	\textbf{Recall} & \ibdtRec & \ivdtRec \\
	\textbf{Precision} & \ibdtPrec & \ivdtPrec \\
	\textbf{Specificity} & \ibdtSpec & \ivdtSpec \\
	\textbf{Accuracy} & \ibdtAcc & \ivdtAcc \\
	\bottomrule
	\end{tabular}
	\caption{
		Average algorithm performance per dataset per subject per
		movement class ($n=4\times6\times3=72$).
	}
	\label{tbl:overview}
\end{table}

\subsection{In-depth Analysis}

\begin{figure}[ht]
	\centering
	\includegraphics[width=\columnwidth]{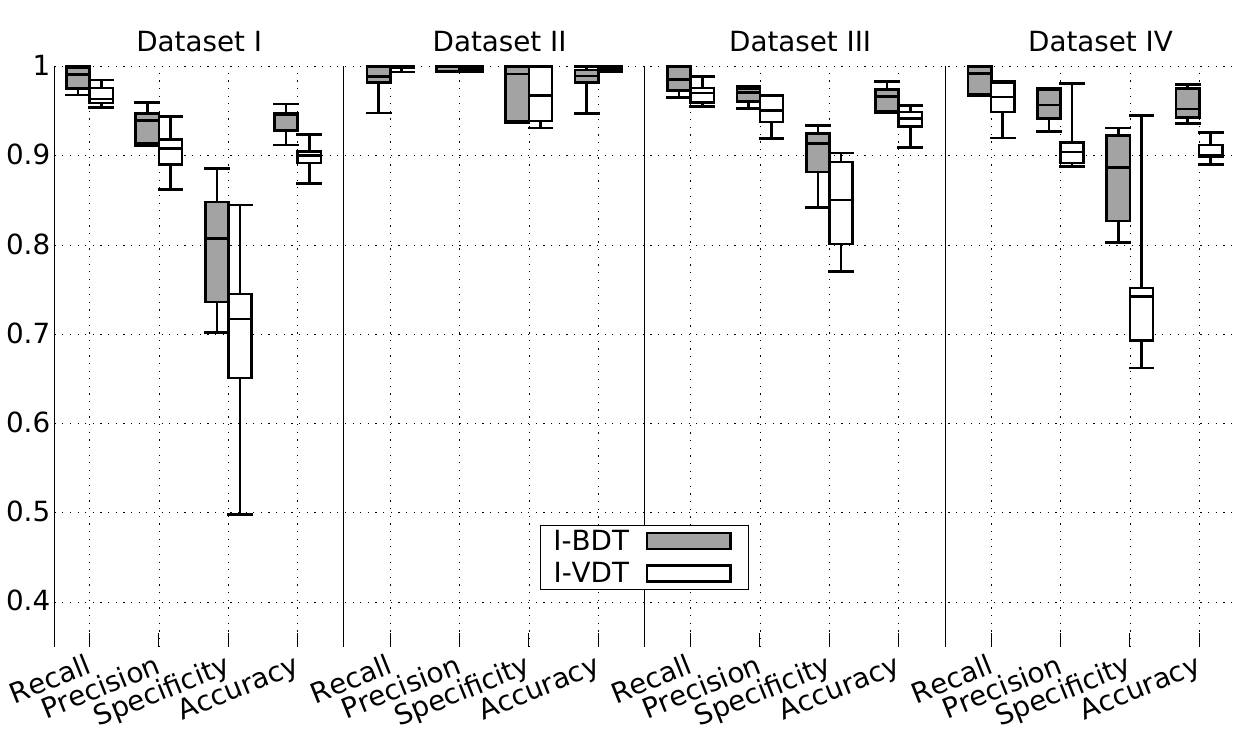}
	\caption{ Performance metrics per dataset for fixations.}
	\label{fig:fix}
\end{figure}

We start our in-depth analysis by looking at the algorithms performance per
dataset for fixations.
\figref{fig:fix} shows that the algorithm scores highly for the recall and
precision metrics for this class, consistently above $90\%$, and generally above
$95\%$.
However, since fixations are the prevalent class in all datasets, false
positives are drowned in the larger number of true positives; as a result, it is
of great importance to look at the specificity when evaluating fixation
classification performance.
In this case, I-BDT scored above $80\%$ reliably.
It is plain that the specificity for dataset~II is well above the others, which
suggests that the false positives are mostly misclassified smooth pursuits.
This is supported by evidence that slow smooth pursuits are the ones being
misclassified; specificity for dataset I is almost consistently lower than for
dataset~III~and~IV, presumably due to dataset I always including the slowest
smooth pursuits. Likewise, specificity for dataset~IV is only sometimes lower
than that of dataset~III because dataset~IV only randomly includes the slowest
smooth pursuits.

\begin{figure}[ht]
	\centering
	\includegraphics[width=\columnwidth]{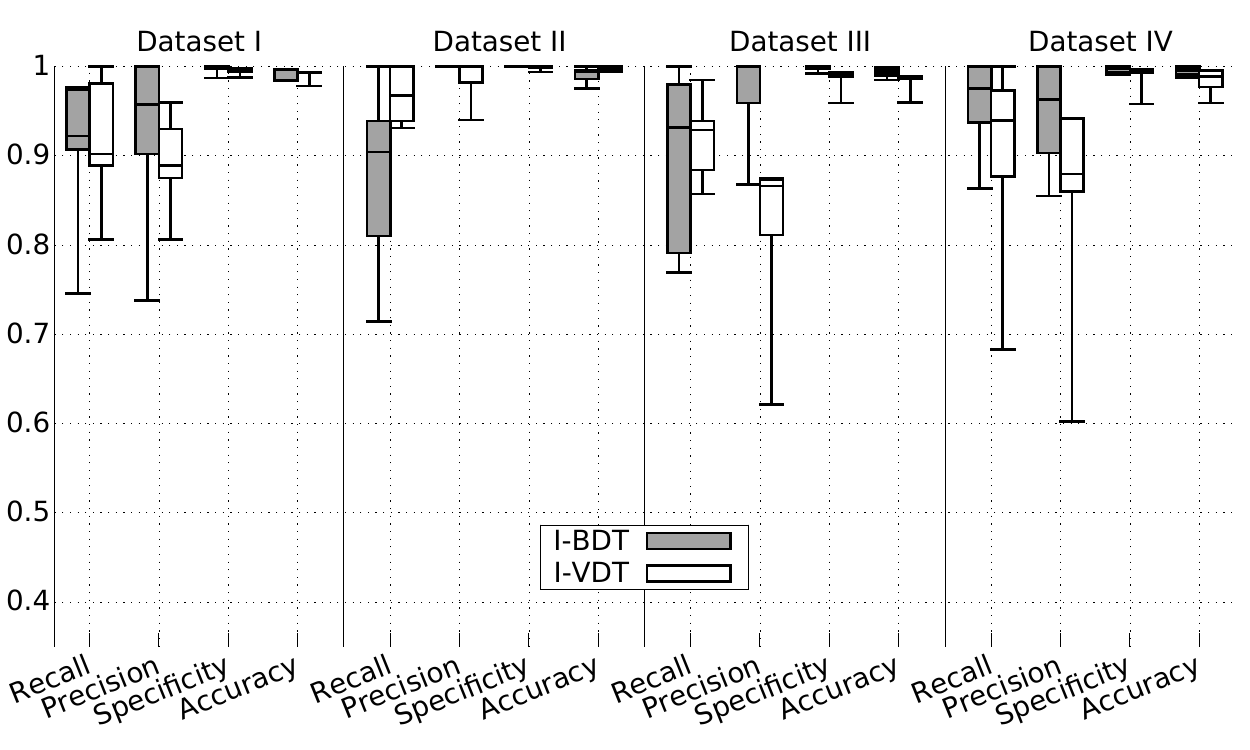}
	\caption{ Performance metrics per dataset for saccades.}
	\label{fig:sac}
\end{figure}

\begin{figure}[ht]
	\centering
	\includegraphics[width=\columnwidth]{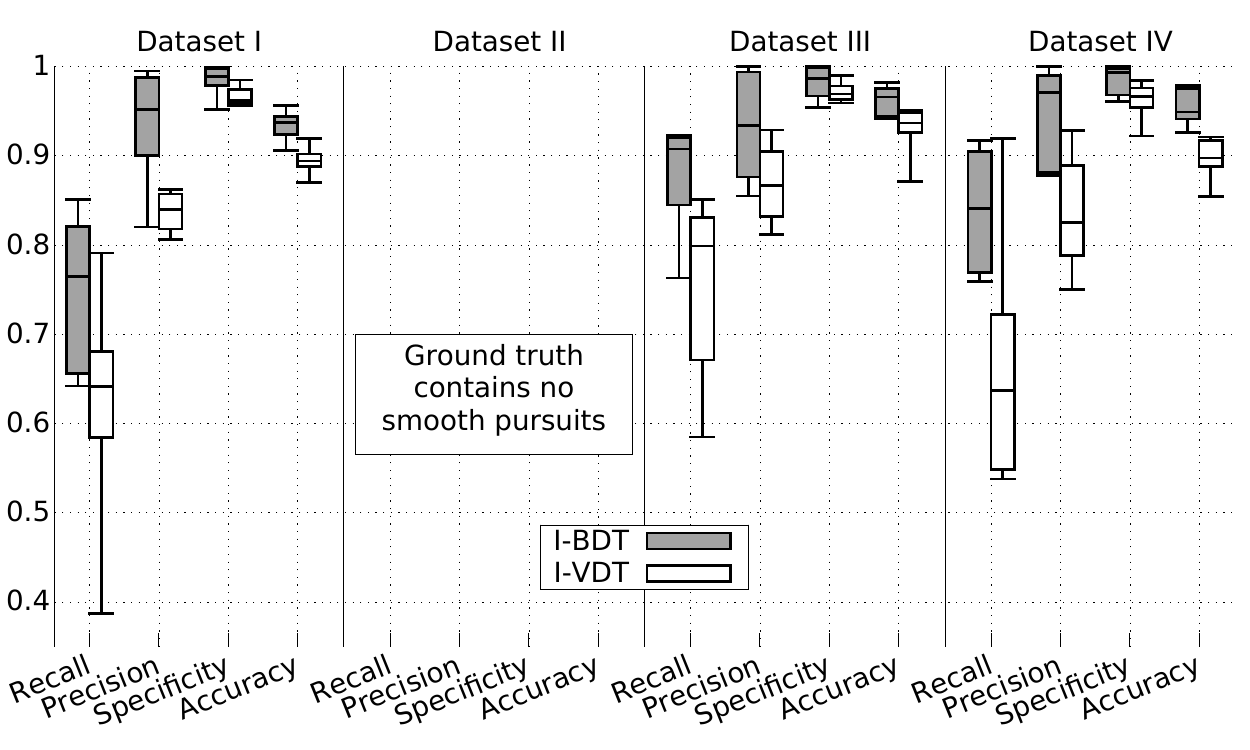}
	\caption{
		Performance metrics per dataset for smooth pursuits. Dataset~II
		contains no smooth pursuits in the ground truth; thus, the resulting
		performance metrics are irrelevant and not reported.
	}
	\label{fig:pur}
\end{figure}

As can be seen in \figref{fig:sac} and \figref{fig:pur}, specificity for both
saccades and smooth pursuits classification is persistently high ($>95\%$).
However, similarly to how precision can be misleading for the performance
evaluation of fixation classification, specificity can be deceptive for saccades
and smooth pursuits classification as false positives get masked by the larger
amount of true negatives. Thus, we analyze saccade and smooth pursuit
classification through the recall and precision metrics.

\figref{fig:sac} shows that saccade classification is very precise ($>90\%$) in
the majority of cases. While the proposed algorithm also displayed a good recall
(mostly above $80\%$), it is clear that some saccades are being misclassified;
these are usually saccades surrounded by noise, which the algorithm ends up
interpreting as a high movement ratio and, thus, classifying as smooth pursuits.
This effect also leads to the I-VDT algorithm outperforming I-BDT for saccade
recall for dataset~II.
Since this dataset contains no smooth pursuits, there is a clear velocity
threshold separating the remaining movements, and, thus, I-VDT can clearly
distinguish between them.
I-BDT, however, is still affected by saccades surrounded by noise, on average
classifying \ibdtFpur{} of the samples as smooth pursuits.
In contrast, dataset~III exposes one of the I-VDT weaknesses as it contains
smooth pursuits with higher speeds (i.e., \SI{44}{\degree/\second}); as a
result, smooth pursuit and saccade speeds overlap, yielding the misclassification
of some high-speed pursuits and decreasing saccade classification precision.

Regarding smooth pursuit classification performance, \figref{fig:fix} highlights
the consistent good precision ($>80\%$) through all datasets, scoring above $90\%$
in the great majority of cases. Analyzing recall performance, I-BDT exhibits good
recall ($>85\%$) for datasets III and IV. As mentioned previously, for dataset I
there is a struggle to classify slow smooth pursuits, resulting in the smaller
recall for this dataset.
Furthermore, it is worth noticing that I-BDT cannot reach maximum recall by
design; since the algorithm relies on a temporal window to consider smooth
pursuits, there is an onset period after the smooth pursuit has started until
I-BDT starts classifying samples as such.

\begin{figure}[ht]
	\centering
	\includegraphics[width=\columnwidth]{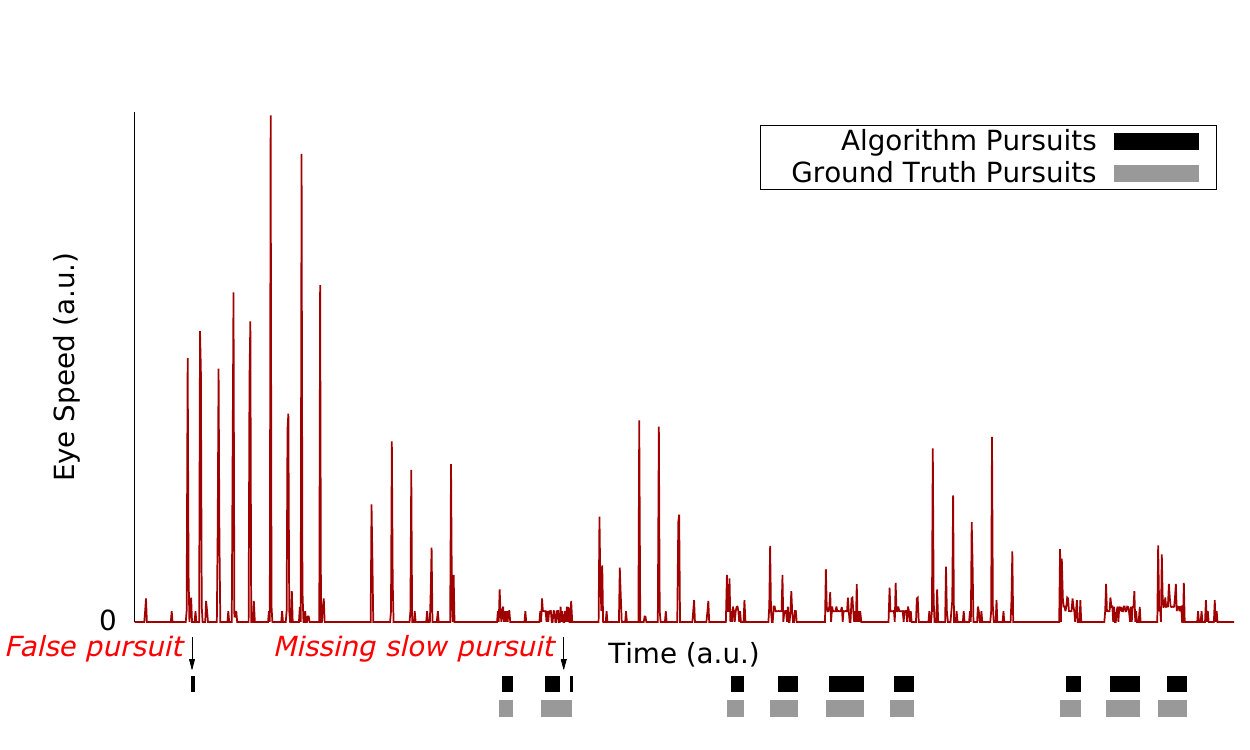}
	\caption{
		I-BDT smooth pursuit classification compared to that of a domain expert,
		accompanied by the eye-speed signal.
		A wrongly detected smooth pursuit and a partially detected slow smooth
		pursuit are highlighted.
		Moreover, notice the onset period required by the algorithm to classify
		the smooth pursuits.
	}
	\label{fig:improvements}
\end{figure}

\figref{fig:improvements} illustrates I-BDT's smooth pursuit classification
relative to that of a domain expert.
Notice how the algorithm detects a false short smooth pursuit sequence at the
beginning due to a saccade surrounded by noise.
In an offline version, such misclassifications could be eliminated, for example,
by using a minimum duration threshold for smooth pursuits; the one in
question, has a duration of approximately only \SI{100}{\milli\second}.
Moreover, it is possible to perceive the onset period for the smooth pursuit
detection at the beginning of each smooth pursuit; this onset period could also
be dealt with in an offline version by employing a similar detection technique
but reversing the order of the samples.
Furthermore, notice that during the second smooth pursuit the eye speed quickly
switches between zero and close to zero values, misleading the algorithm, which
does not detect the whole slow pursuit successfully.
Thus, we do not advise the usage of I-BDT as is for very slow smooth pursuits
when using low-resolution eye trackers; higher resolutions should alleviated
this problem, but further investigation is required.
It is worth noticing that, despite this weakness, low-resolution eye trackers
are more appealing for embedded use in dynamic scenarios because these systems
are cheaper, less computationally intensive, and consume less power than
their high-resolution counterparts.

Comparing our results to those of related work is relatively complicated, mainly
due to the lack of openness regarding algorithms and datasets, and also due to
differences in eye-tracking systems and metrics used for evaluation.
Regarding dataset, eye-tracking system, and online constraints, the work
conducted by~\cite{vidal2012detection} is the most similar to the one from this
work.
In fact, our dataset design was heavily influenced by the dataset used in that
work (better described in~\cite{vidal2011analysing}) and the one
from~\cite{larsson2013detection}; the main differences are 1) smooth pursuits in
this work are not restricted to horizontal and vertical directions, and 2) we
chose not to include short smooth pursuits (e.g., amplitude of
\SI{2}{\degree} and velocity of \SI{30}{\degree/\second}) as their durations are
smaller than an acceptable latency for the subject to start tracking the target.

In their work, \cite{vidal2012detection} report an accuracy for smooth pursuit
detection up to $92\%$ while, in this work, I-BDT reached an average accuracy of
\ibdtAvgPurAcc{}, ranging from \ibdtMinPurAcc{} to \ibdtMaxPurAcc{}.
It is worth noticing that accuracy alone does not allow us to completely
evaluate algorithm performance~\cite{ben2007lot}.
Unfortunately, the machine learning-based classifier presented in
\cite{vidal2012detection} is not available for evaluation on our dataset, nor is
their dataset available for evaluation with other algorithms.
Thus, a direct comparison of both methods could not be performed.

\cite{larsson2015detection} use a subset of the dataset
from~\cite{larsson2013detection}; however, their algorithm is designed for
offline analysis of high-resolution eye-tracking data.
Thus, it cannot be applied to low-resolution eye trackers such as the one used
in this work -- mainly due to the preliminary segmentation stage relying on
hypothesis testing, which would require a long time interval from the
low-resolution eye tracker to be statistically significant
(\SI{363}{\milli\second} compared to the \SI{22}{\milli\second} used in their
work).
Nonetheless, their static \emph{image} dataset can to some extent be compared to
dataset~II (in the sense that both do not contain smooth pursuits inducing
elements). Similarly, their \emph{video} and \emph{moving dot} datasets can be
compared to datasets~I,~III,~and~IV.
Since their algorithm uses the same mechanism as I-VDT to separate saccades from
other eye movements, their algorithm performance in this regard is clear.
Thus, we briefly draw a parallel between their results for smooth pursuit and
fixation classification and our results.
\tblref{tbl:larsson} reports recall (i.e., sensitivity) and specificity values
from I-BDT mean results from this work, as well as best case results
from~\cite{larsson2015detection} -- to pick a best case scenario, we utilize the
maximum value independent from which expert (1 or 2) was used as ground truth.
Although I-BDT seems to provide better performance despite working under harder
constraints, a fair and valid conclusion could only be drawn from similar
experiments. Nonetheless, it is worth noticing that such an experiment is possible
as I-BDT could be applied to the datasets from~\cite{larsson2015detection}
(e.g., by coalizing the data into a lower resolution or applying I-BDT with
adapted parameters).
Unfortunately, neither dataset nor algorithm implementation
from~\cite{larsson2015detection} are openly available.
\begin{table}[ht]
	\small
	\centering
	\begin{tabular}{llcccc}
	\toprule

		  &          & \multicolumn{2}{c}{\textbf{Recall}} & \multicolumn{2}{c}{\textbf{Specificity}} \\
		  &          & \textbf{I-BDT}                      & \textbf{Larsson}                            & \textbf{I-BDT} & \textbf{Larsson}  \\
\midrule
\multirow{2}{*}{Static}  & Fixation & \ibdtFixSRec                        & $\approx$0.93                               & \ibdtFixSSpec  & $\approx$0.98\\
		  & Pursuit  & \ibdtPurSRec                        & $\approx$0.75                               & \ibdtPurSSpec  & $\approx$0.97\\
\midrule
\multirow{2}{*}{Dynamic} & Fixation & \ibdtFixDRec                        & $\approx$0.90                               & \ibdtFixDSpec  & $\approx$0.85\\
		  & Pursuit  & \ibdtPurDRec                        & $\approx$0.80                               & \ibdtPurDSpec  & $\approx$0.95\\

	\bottomrule
	\end{tabular}
	\caption{
		Performance comparison between I-BDT and~\protect\cite{larsson2015detection}.
		Static represents the average performance for dataset~II compared to the
		best performance for the images dataset.
		Dynamic represents the average performance for datasets~I,~III,~and~IV compared to the
		best performance for the videos/moving dot datasets.
	}
	\label{tbl:larsson}
\end{table}


\section{Final Remarks}

In this paper, we have proposed and evaluated a novel algorithm for the
real-time identification of fixations, saccades, and smooth pursuits for
low-resolution eye trackers.
Since the algorithm operates directly on the eye-position signal, it requires no
calibration step.
The proposed algorithm displayed higher and more consistent performance than an
state-of-the-art algorithm, demonstrating the capability of I-BDT to provide
meaningful ternary classification.
Moreover, an open-source \emph{MATLAB} implementation of the
algorithm is provided.

One of the main difficulties during evaluation, was the lack of open annotated
datasets.
The manual coding of eye movements is a subjective, laborious, and
time-consuming task; thus, having to create one from scratch is far from ideal.
In an effort to allow for review and ease the evaluation of eye movement
identification algorithms, we are willing to provide our annotated datasets;
please contact the first author in order to obtain them.

For future work, we are interested in analyzing additional features for I-BDT to
further improve its performance, as well as evaluating the algorithm with
higher-resolution eye trackers.
Moreover, an important step to enable the fully automation of eye movements
classification is a reliable detection of blinks, which the proposed algorithm
does not take into account at the moment.
Furthermore, we intent on developing solutions to account for head movements in
order to reliably distinguish smooth pursuits from vestibulo-ocular
reflexes.

\bibliographystyle{acmsiggraph}
\bibliography{references}

\end{document}